\documentclass[11pt,a4paper]{article}
\usepackage[hyperref]{acl2019}
\usepackage{times}
\usepackage{latexsym}

\usepackage{verbatim}
\usepackage{url}
\usepackage{textcomp}
\usepackage{graphicx}
\usepackage{multicol}
\usepackage{booktabs}
\usepackage[linesnumbered,algoruled,boxed,lined]{algorithm2e}

\graphicspath{ {./figures/} }
\usepackage{amsmath}

\aclfinalcopy 
\def\aclpaperid{2716} 

\title{Eliciting Knowledge from Experts:
\\
Automatic Transcript Parsing for Cognitive Task Analysis}

\author{Junyi Du\textsuperscript{1}, He Jiang\textsuperscript{1}, Jiaming Shen\textsuperscript{2}, Xiang Ren\textsuperscript{1} \\
  \textsuperscript{1}Department of Computer Science, University of Southern California \\ \textsuperscript{2}Department of Computer Science, University of Illinois at Urbana-Champaign \\
  \texttt{\{junyidu, jian567, xiangren\}@usc.edu, js2@illinois.edu}}

\date{}

\newcommand{\calL}{\mathcal{L}}

\newcommand{\calM}{\mathcal{M}}
\newcommand{\calS}{\mathcal{S}}
\newcommand{\bp}{\mathbf{p}}
\newcommand{\bt}{\mathbf{t}}
\newcommand{\bx}{\mathbf{x}}

\newcommand{\bh}{\mathbf{h}}
\newcommand{\bu}{\mathbf{u}}
\newcommand{\bv}{\mathbf{v}}

\begin{document}
\maketitle
\begin{abstract}
Cognitive task analysis (CTA) is a type of analysis in applied psychology aimed at eliciting and representing the knowledge and thought processes of domain experts. In CTA, often heavy human labor is involved to parse the interview transcript into structured knowledge (e.g., flowchart for different actions). To reduce human efforts and scale the process, automated CTA transcript parsing is desirable. However, this task has unique challenges as (1) it requires the understanding of long-range context information in conversational text; and (2) the amount of labeled data is limited and indirect---i.e., context-aware, noisy, and low-resource.
In this paper, we propose a weakly-supervised information extraction framework for automated CTA transcript parsing. We partition the parsing process into a sequence labeling task and a text span-pair relation extraction task, with distant supervision from human-curated protocol files.
To model long-range context information for extracting sentence relations, neighbor sentences are involved as a part of input. Different types of models for capturing context dependency are then applied. We manually annotate real-world CTA transcripts to facilitate the evaluation of the parsing tasks\footnote{\scriptsize Code is available at: \href{https://github.com/cnrpman/procedural-extraction}{https://github.com/cnrpman/procedural-extraction}}.
\end{abstract}

\begin{figure}[hbt!]
 \centering
 \includegraphics[width=\linewidth]{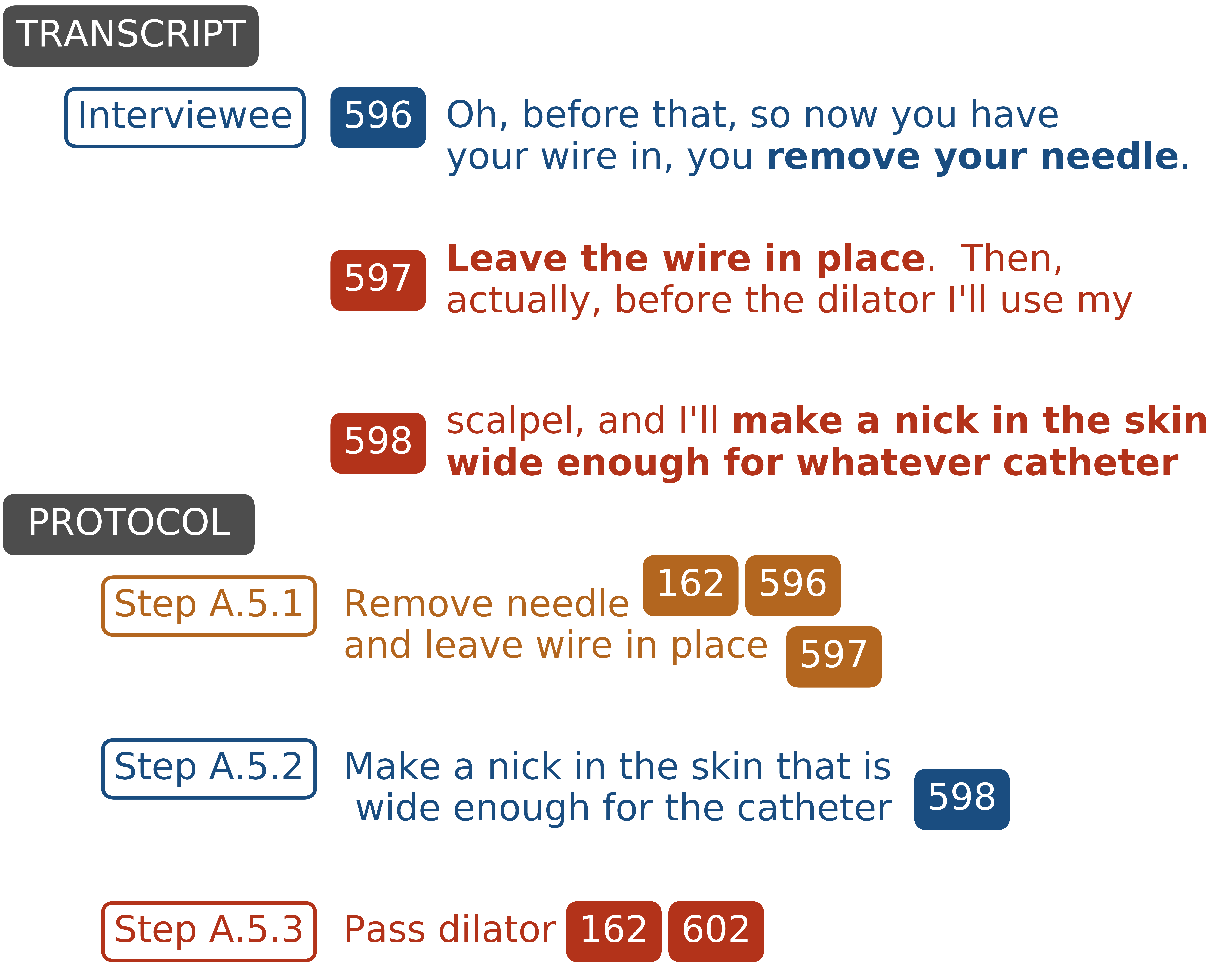}
 \caption{\textbf{An example of CTA interview transcript and the human parsed structured text (protocol).} In the protocol, splitting by the highlighted line numbers indicating the sources in transcript, phrases in protocol (called \textit{protocol phrases}) are abstractive description of actions in the transcript. In the transcript, the highlighted numbers are line numbers, and the bolded are \textit{text spans} matched by protocol phrases. The highlighted line numbers are provided by human parsing which provide constraint on mapping protocol phrases back to the transcript, but they are noisy and pointing back to a large scope of sentences, instead of the text span we want to extract.}
  \label{fig:figure1}
\end{figure}

\begin{figure*}[hbt!]
 \centering
 \includegraphics[width=\textwidth]{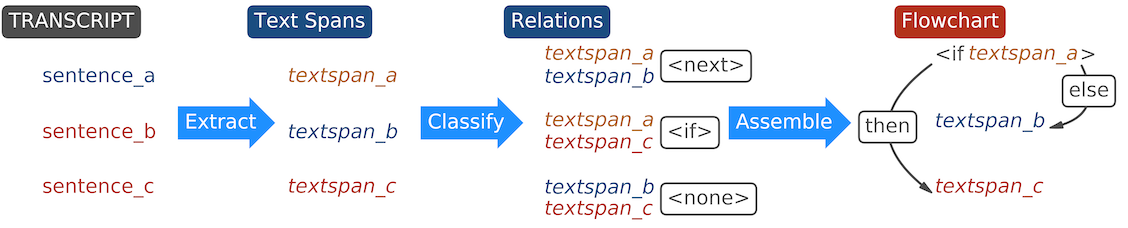}
 \caption{\textbf{The framework of Automated CTA Transcripts Parsing.} Text spans are extracted via the sequence labeling model, then the relations between text spans are extracted by the text span-pair relation extraction model (\textit{span-pair RE} model). In the end we assemble the results into structured knowledge (flowchart) for CTA.}
 \label{fig:predict}
\end{figure*}

\section{Introduction}
\textit{Cognitive task analysis} (CTA) is a powerful tool for training, instructional design, and development of expert systems \cite{woods1989cognitive,clark1996cognitive} focusing on yielding the knowledge and thought processes from domain experts \cite{schraagen2000cognitive}. Traditional CTA methods require interviews with domain experts and parsing the interview transcript (\textit{transcript}) into structured text describing processes (\textit{protocol}, shown in Fig. \ref{fig:figure1}). However, parsing transcripts requires heavy human labor, which becomes the major hurdle of scaling up CTA. Therefore, automated approaches to extract structured knowledge from CTA interview transcripts are important for expert systems using massive procedural data.

A natural realization of automated CTA is to apply relation extraction (RE) models to parse interview text. However, the key challenge here is the lack of direct sentence-level supervision data for training RE models because the only available supervision, protocols, are document-level transcripts summaries. Furthermore, the information towards relations between procedural actions spreads all over the transcripts, which burdens the RE model to process global information of the text. One previous work \cite{codifyhowto} studies extracting procedure information on \emph{well-structured text} using \verb|OpenIE| and sentence pair RE models. In this work, however, we focus on \emph{unstructured conversational text} (i.e., CTA interview transcripts) for which \verb|OpenIE| is inapplicable.

To address the above challenges, we develop a novel method to effectively extract and leverage weak(in-direct) supervision signals from protocols. The key observation is that these protocols are structured in the phrase level (c.f. Fig.~\ref{fig:figure1}). We split each protocol into a set of \textit{protocol phrases}. Each protocol phrase is associated with a line number that points back to one sentence in the original transcript. Then, we can map these protocol phrases back to \textit{text spans} in transcript sentences and obtain useful supervision signals from three aspects. First, these matched text spans provide direct supervision labels for training text span extraction model. Second, the procedural relations between protocols phrases are transformed into relations between text spans within sentences, which enables us to train RE models. Finally, the local contexts around text spans provide strong signals and can enhance the mention representation in all RE models.

Our approach consists of following steps: (1) parse original protocol into a collection of protocol phrases together with their procedural relations, using a deterministic finite automation (DFA); (2) Match the protocol phrases back to the text spans in transcripts using \textit{fuzzy matching} \cite{pennington2014glove, bert}; (3) Generate text span extraction dataset and train a sequence labeling model \cite{crfner, lmlstmcrf} for text span extraction; (4) Generate text span-pair relation extraction (span-pair RE) dataset and fine-tune pre-trained context-aware span-pair RE model~\cite{bert}. With the trained models, we can automatically extract text spans summarizing actions from transcripts along with the procedural relations among them. Finally, we assemble the results into protocol knowledge, which lays the foundation for CTA.

We explore our approaches from manifold aspects: (i) We experimented different fuzzy matching methods, relation extraction models and sequence labeling models; (ii) We present models for solving context-aware span-pair RE; (iii) We evaluate the approach on real-world data with human annotations, which demonstrates the best fuzzy matching method achieves 47.1\% mention level accuracy, best sequence labeling model achieves 38.18\% token level accuracy, and best text span-pair relation extraction model achieves 74.4\% micro F\textsubscript{1}.
\section{Related Work}
Our work is closely related to procedural extraction, however we focus on conversational text from CTA interviews which is in a low-resource setting and no sentence-by-sentence label is available.

\noindent \textbf{Cognitive task analysis.}
Cognitive task analysis is a powerful tool for extracting knowledge and thought processes of experts widely used in different domains \cite{schraagen2000cognitive, seamster2017applied}. Yet, it is time-consuming and not scalable. Recent years, with the development of natural language processing, techniques are introduced to aid human expertise \cite{Zhong:2015,etcta}. \citeauthor{li2013general}\shortcite{li2013general} used learning agent to discover cognitive model in specific domains. \citeauthor{cmwithnn}\shortcite{cmwithnn} explored modeling cognitive knowledge in well-defined tasks with neural models. However, for the most general setting that extract cognitive processes from interviews, we still need substantial expertise to interpret the interview transcript.

\noindent \textbf{Procedural extraction.}
Recent advances in machine reading comprehension, textual entailment~\cite{bert} and relation extraction \cite{zhang2017position} shows the contemporary NLP models have the capability of capturing causal relations in some degree. However, it is still an open problem to extract procedural information from text. There were some attempts to extract similar procedural information on well-structured instructional text from how-to community. \citeauthor{codifyhowto} \shortcite{codifyhowto} treated procedural extraction as a relation extraction problem on sentence pair extracted by pattern matching. They used \verb|OpenIE| for pattern extraction and hierarchical LSTM to classify relation labels of sentence pairs.

\noindent \textbf{Pre-trained language representations.}
Recent researches showed that language models generically trained on massive corpus is beneficial to various specific NLP tasks~\cite{pennington2014glove, bert}. Language representation has been an active area of research for years. Tons of effective approaches have been developed from feature-based approaches \cite{ando2005framework, mikolov2013distributed, elmo2018} to fine-tuning approaches \cite{dai2015semi,openaigpt, bert}. 
\begin{figure*}[hbt!]
 \centering
 \includegraphics[width=\textwidth]{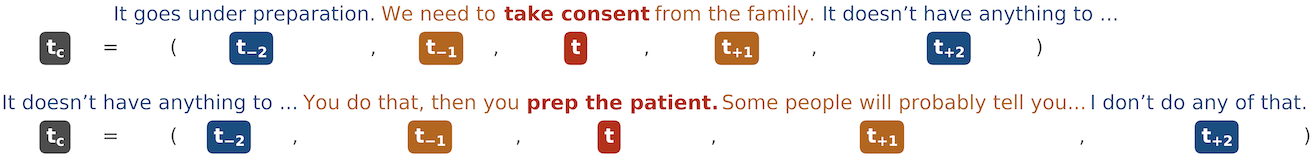}
 \caption{\textbf{The construction of text span with context $\mathbf{t}_c$.} The example shows two text spans with context using $K=2$. Neighbours of text span $\mathbf{t}$ are denoted by $\mathbf{t}_{+i}$ and $\mathbf{t}_{-i}$, $0 < i <= K$}
 \label{fig:context}
\end{figure*}
\section{Framework}
Our automated CTA transcript parsing framework takes interview transcripts as input and outputs structured knowledge consisting of summary phrases. The framework, visualized in Fig.~\ref{fig:predict}, includes two parts: (1) summary text spans extraction and (2) text span-pair relation extraction. The extracted knowledge will then be structured using a flowchart and supports automated CTA.

\subsection{Text Spans Extraction}\label{sec:tse}
Since CTA interview transcripts are conversational text while structured knowledge are formed of summary phrases describing actions in transcripts (c.f. Fig.~\ref{fig:figure1}), we need to first summarize transcript sentences. An intuitive idea is to first leverage off-the-shelf text summarization methods \cite{shen2007document, rnnsum, AAAIgansum}. However, CTA is a low-resource task and thus we do not have enough training data for learning seq2seq-based text summarization models. Therefore, in this work, we formulate the summarization of transcript sentences as a sequence labeling task~\cite{lmlstmcrf} and treat the best summarized text span in a transcript sentence as its corresponding summary phrase.

Given a sentence in transcripts, we denote the sentence as $\bx=\{x_i\}$ where $x_i$ is the token at position $i$. The text spans extraction task aims to obtain the prediction $\bp_t$ representing the summary text span $\bt$ of the transcript sentence $\bx$ using a sequence labeling model $\bp_t = \calM_{s}(\bx)$, where $\bt$ is a continuous subset of $\bx$ labeled by $\bp_t=\{{p_t}_i\}$ with \verb|IOBES| schema. To train the model, we utilize weakly-supervised sequence labels created in Sec. \ref{subsec:seqlabel}.

\subsection{Text Span-Pair Relation Extraction}\label{sec:tsre}
Structural relations between text spans are required to assemble summary text spans into structured knowledge. To extract structural information, following the previous study \cite{codifyhowto}, we formalize text span-pair relation extraction as a sentence pair classification problem. 
A directed graph $\mathcal{G}_t=(\mathcal{T}, \mathcal{R}_t)$ is used to represent the structured knowledge parsed from a CTA transcript, consisting of nodes for summary text spans in the transcript ($\mathcal{T}=\{\mathbf{t}_i\}$) and edges for procedural information ($\mathcal{R}_t=\{({\mathbf{u}_t}_i,{\mathbf{v}_t}_i,{r_t}_i)\}$ where ${\mathbf{u}_t}_i, {\mathbf{v}_t}_i \in \mathcal{T}$ are summary text spans and ${r_t}_i$ is the procedural relation from text span ${\mathbf{u}_t}_i$ to text span ${\mathbf{v}_t}_i$). 
A span-pair RE model ${r_t}_i = \mathcal{M}_r({\mathbf{u}_t}_i, {\mathbf{v}_t}_i), \forall {\mathbf{u}_t}_i, {\mathbf{v}_t}_i \in \mathcal{T}$ is then applied to extract relations between all summary text spans $\mathcal{T}$ in the transcript.
We train the model using the span-pair RE dataset generated in Sec.~\ref{subsec:relation}.

To capture the long-range context dependency, we enrich the text span representation $\mathbf{t}$ based on its surrounding contexts and feed the enhance span representation $\mathbf{t}_c$ into the relation extraction model $\mathcal{M}_r$. Examples are shown in Fig.~\ref{fig:context}.

\subsection{Context-aware Models for Text Span-Pair Relation Extraction}
\label{sec:context}
We apply state-of-the-art models for natural language entailment~\cite{hbmp, bert} to solve the text span-pair relation extraction task as a sentence pair classification problem. 
While these models show promising results on the span-pair RE dataset we generated, they do not fully exploit all the information of our dataset. For example, in our dataset, a text span with context information is the combination of matched text span and its surrounding context sentences (Fig.\ref{fig:context}). But in the normal sentence pair classification setting, they are concatenated into a single sequence while its segmentation is ignored. Here, we explored some variants of the neural model, to incorporate the context segmentation and position information with the state-of-the-art model for sentence pair classification.

 \vspace{0.2cm}
\begin{figure}[hbt!]
 \centering
 \includegraphics[width=\linewidth]{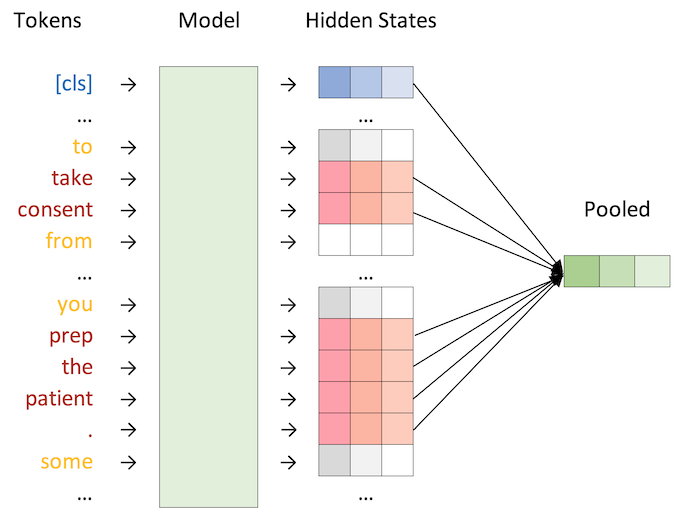}
 \caption{\textbf{Visualization of the hidden state masking.} Hidden states for the context sentences are masked before pooling.}
\vspace{0.2cm}
 \label{fig:mask}
\end{figure}

\medskip
\noindent \textbf{Hidden states masking.}
In this model variant we inject the context segmentation into models by masking out the final layer hidden states for the context sentences and aggregate the remaining hidden states using a pooling function. This structure enables us to incorporate context segmentation information without introducing any new parameters.
 \vspace{0.1cm}
\begin{align}
    H_{\bt} & = \{\bh_i | t_i \in \bt\} \notag \\
    \bh_{\text{MAX}} &= \max(H_{\bu_t} \cup H_{\bv_t} \cup h_{[cls]} \cup h_{[sep]})\\
    \bh_{\text{AVG}} &= \frac{\sum{H_{\bu_t}} + \sum{H_{\bv_t}} + h_{[cls]} + h_{[sep]}}{|\bu_t| + |\bv_t| + 2}
\end{align}
 \vspace{0.1cm}
where $\{\bh\}$ are the final layer hidden states, $\bu_t$, $\bv_t$ are the two tokenized text spans, $t_{[cls]}, t_{[sep]}$ are the $[cls]$ token and $[sep]$ token (for BERT model), $h_{[cls]}, h_{[sep]}$ are the corresponding hidden states, respectively.

\medskip
\noindent \textbf{Import context position as attention.}\label{sec:contextseq}
Inspired by position embedding and position-aware attention \cite{posemb, zhang2017position}, we define two context position sequences $[c_1,\cdots,c_n]$ and $[c'_1,\cdots,c'_n]$, which correspond to the position of the two text spans, respectively, that is:
 \vspace{0.1cm}
\begin{equation}  
c_i=\left\{
\begin{aligned}
p_i - p_t - 1, ~~~p_i < p_t \\
1\textrm{ or }-1, ~~~p_i = p_t \\
p_i - p_t + 1, ~~~p_i > p_t
\end{aligned}
\right.
\end{equation}
\vspace{0.1cm}
We use $p_i$ and $p_t$ to denote the position of context and text span in transcript in sentence level. For $i = p_t$, $c_i = 1$ or $-1$, depends on whether the context is on the left or right of the text span. The two context position sequences are truncated by a fix length for computational complexity, then injected into BERT model using position-aware attention \cite{zhang2017position}.

\medskip
\noindent \textbf{Import context position as input embedding.}
Segment embedding is a part of input embedding designed to import sentence-pair segmentation information in BERT model. In this model variant we expand the segment embedding to encode context position sequence above.
\begin{figure}[hbt!]
 \vspace{0.6cm}
 \centering
 \includegraphics[width=1.01\linewidth]{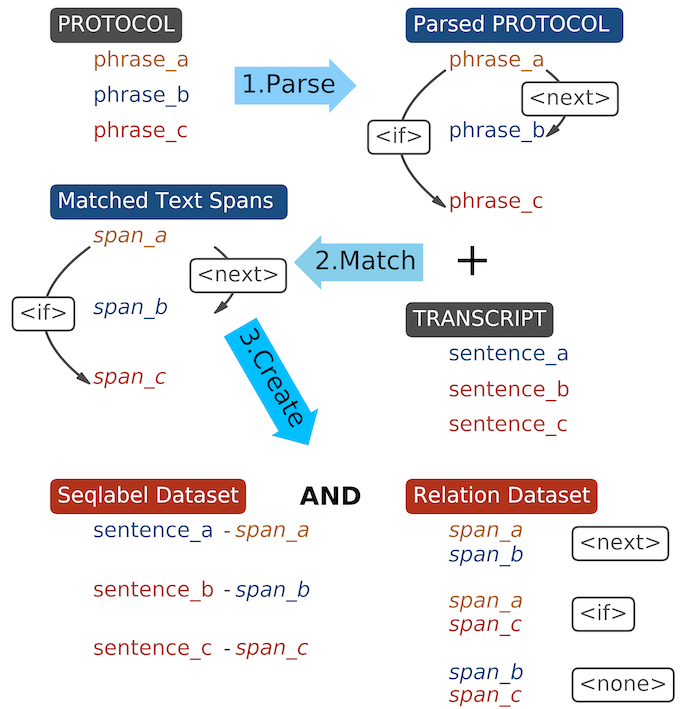}
 \caption{\textbf{The dataset generation pipeline.} The protocol is first parsed into a graph with relations between protocol phrases (shown as \textit{phrase}), then match the protocol phrases with the text spans in transcripts (shown as \textit{span}). Finally, sequence labeling dataset and span-pair RE dataset are created according to the matches and the relations.}
\vspace{0.2cm}
 \label{fig:pipeline}
\end{figure}

\vspace{0.2cm}
\section{Dataset Generation}
To take advantage of the weak supervision from protocols, we build a pipeline to generate datasets for the CTA parsing framework, showed in Fig.~\ref{fig:pipeline}.
\subsection{Protocol Parsing}
We use a deterministic finite automation to parse the protocol into a graph $\mathcal{G}_p=(\mathcal{P},\mathcal{R}_p)$ describing protocol phrases represented by nodes ($\mathcal{P}=\{\mathbf{p}_i\}$ which denotes all protocol phrases parsed from the protocol) and procedural relations represented by edges ($\mathcal{R}_p=\{({\mathbf{u}_p}_i,{\mathbf{v}_p}_i,{r_p}_i)\}$, where ${\mathbf{u}_p}_i, {\mathbf{v}_p}_i \in \mathcal{P}$ are protocol phrases and ${r_p}_i$ is the procedural relation from phrase ${\mathbf{u}_p}_i$ to phrase ${\mathbf{v}_p}_i$).

We consider three types of procedural relations during the parsing: $\langle$none$\rangle$ for no procedural relation between protocol phrases, $\langle$next$\rangle$ for sequence, and $\langle$if$\rangle$ for decision branching.

\subsection{Text Spans Matching}
To enable the abundant information in protocols, we want to map each phrase in the protocol back to the nearest textual representation in the transcript. We can achieve this by using sentence matching techniques. Following the sequence labeling setting in transcript summarization of our framework, given a protocol phrase $\mathbf{p}$, we want to find the best matching text span $\mathbf{t}$ in the transcript. The scope of search is limited to the source lines $\mathcal{L}_{\mathbf{p}}$ mentioned in the protocol (Fig.~\ref{fig:figure1}). Then we extract all possible text spans $\{\mathbf{t}_i\}$ from these sentences by enumerating all available $n$-grams and find the best matching text span $\mathbf{t}_{\text{best}}$ for $\mathbf{p}$ that maximizes sentence similarity measure $\mathcal{M}_{sim}(\mathbf{p}, \mathbf{t}_{\text{best}})$. Following is the overall workflow:
\begin{align}
    \calS & = \{\text{retrieve\_sentence}(\ell) | \ell \in \calL_{\mathbf{p}}\} \notag \\
    \bt_{\text{best}} & = \arg \max_{\bt \in \calS} \calM_{\text{sim}} (\bp, \bt) \notag \\
    \calM_{\text{sim best}} & = \max_{\bt \in \calS} \calM_{\text{sim}} (\bp, \bt) \\
    \text{match} & = \begin{cases}
        \text{None} & \calM_{\text{sim best}} \leq \text{threshold}\\
        \bt_{\text{best}} & \calM_{\text{sim best}} > \text{threshold}
    \end{cases} \notag
\end{align}

For the similarity measure $\mathcal{M}_{\text{sim}}$, we adopt sentence embedding from different methods~\cite{pennington2014glove, bert}. The similarity is calculated by the cosine distance between two normalized sentence embedding. An empirical threshold 0.5 is adopted for dropping the protocol phrases without good matched text span. We then match the protocol phrases back to the nearest text span in the transcript.

\subsection{Sequence Labeling Dataset}\label{subsec:seqlabel}
With the matched text spans in the transcript, we are able to assign labels to every token in the transcript, denoting whether the token belongs to a matched text span. We adopt IOBES format \cite{iob} as the labeling schema for constructing the sequence labeling dataset. The labeled text spans are semantically close to the protocol phrases which are abstractive description of actions, and we can use the labels to train text spans extraction models (Sec.~\ref{sec:tse}) in a weakly-supervised manner.

\subsection{Text Span-Pair Relation Extraction Dataset}\label{subsec:relation}
By parsing the protocol we learn the procedural relations between protocol phrases. Thus we can apply them to the matched text spans in transcript to construct the span-pair RE dataset. These relations serve as weak supervision for the span-pair RE model (Sec.~\ref{sec:tsre}). Corresponding to the relation types parsed from the protocols, the dataset include three types of label: $<$none$>$, $<$next$>$ and $<$if$>$.

\subsection{Human-Annotated Matching Test Set} \label{sec:manual}
Since the datasets for CTA transcript parsing framework are created via matching, we need to evaluate the performance of our matching methods. Thus, for testing purpose, we manually annotated the matched text spans in transcript for 138 protocol phrases as the manual matching test set. Furthermore, we create two test sets to evaluate the effectiveness of our approach with the manual matching annotations, which are called manual matching sequence labeling test set and manual matching span-pair RE test set. In comparison, we call the test sets generated via text spans matching as generated sequence labeling test set and generated span-pair RE test set.
\section{Experiments}
In this section we evaluate the effectiveness of our proposed automated CTA transcript parsing framework and the models. Especially, we run three sets of experiments: (1) we evaluate our text spans matching methods with the manual matching test set; (2) we evaluate model performance on the CTA text spans extraction task with the sequence labeling dataset; (3) we evaluate model performance on the CTA span-pair RE task with the RE dataset.

\subsection{Text Spans Matching}
\noindent\textbf{Implementation.} We enumerate all text spans with length $[2, K_t]$ within the sentences in transcripts, where $K_t=30$ for truncating text spans. For text spans matching, we try two sentence encoding methods to extract sentence embeddings: (1) average pooling on Glove word embeddings of words in sentences and text spans \cite{pennington2014glove}; (2) extracting features using pre-trained BERT\textsubscript{BASE} model and sum up the features in the last four layers then average over words in sentences and text spans \cite{bert}. Then, we normalize the embeddings and find the best matching text spans for each protocol phrase based on cosine similarity. We also provide the exact matching as a baseline, which finds the longest transcript text span matched by a text span in protocol phrase.
\begin{table}[hbt!]
\small
\begin{center}
\begin{tabular}{ c | c c c }
\toprule
Encoding & Tok. Acc. & Tok. F\textsubscript{1} & Men. Acc.\\
\midrule
Exact & 70.30 & 9.44 & 2.17 \\
\midrule
Glove-50d & 75.40 & 43.92 & 37.68 \\
Glove-300d & \textbf{76.97} & \textbf{45.12} & 42.03 \\ 
BERT fea. & 75.22 & 37.20 & \textbf{47.10} \\
\bottomrule
\end{tabular}
\caption{\textbf{Matching performance on the manual matching testset with different sentence encoding,} in token level accuracy and mention level accuracy. BERT fea. means using features extracted by BERT model. and Exact is the exact matching baseline}
\label{tab:matching}
\end{center}
\end{table}

\noindent\textbf{Evaluation.}
We evaluate the performance of our text spans matching methods with the manual matching test set by token level metrics and mention level accuracy, where token level metrics are normalized by sentence lengths. Results in Table \ref{tab:matching} show the two methods get acceptable results while the exact matching baseline has a poor performance in comparison. Glove-300d shows better token level accuracy and F\textsubscript{1} score while BERT features have a better mention level accuracy. For cheaper computation, we use Glove-300d as the sentence encoding method of matching for the following sections. Please refer to the appendix for the case study of matching.
\subsection{Text Spans Extraction}
\noindent\textbf{Models.} We conduct the experiments of text spans extraction using off-the-shelf sequence labeling models, including CRF \cite{crfner}, LSTM-CRF \cite{bilstmcrf} and LM-LSTM-CRF \cite{lmlstmcrf}. The models are trained on the sequence labeling dataset generated by text spans matching. For comparison, we also implement a hand-crafted rule extraction baseline with TokensRegex.

\noindent\textbf{LSTM-CRF and LM-LSTM-CRF.}
We use LM-LSTM-CRF\footnote{https://github.com/LiyuanLucasLiu/LM-LSTM-CRF} to conduct our experiments for both models, with the same setting of 2 layers word level LSTM, word level hidden size $H_w = 300$, SGD with 0.045 learning rate and 0.05 learning rate decay, and 0.3 dropout ratio. The major difference between two models is that LM-LSTM-CRF contains an additional char-level structure optimized via language model loss. 
\begin{table}[!hbt]
\small
\begin{center}
\begin{tabular}{ c | c c c c}
\toprule
Model & Tok. Acc. & Men. P & Men. R & Men. F\textsubscript{1}\\
\midrule
Rules & - &12.7 & 34.8 & 18.6\\
\midrule
CRF & 80.7 & 38.5 & 37.9 & \textbf{38.1} \\
LSTM-CRF & 75.9 & 40.4 & 31.8 & 35.6 \\
w/ LM\textsubscript{16\ } & 74.6 & 31.8 & 21.2 & 25.5 \\
w/ LM\textsubscript{64\ } & 74.8 & 33.3 & 18.2 & 23.5 \\
\bottomrule
\end{tabular}
\caption{\textbf{Performance of sequence labeling models}, evaluated on manual matching testset. LM-LSTM-CRF is shown as w/ LM, with different character level hidden size.}
\label{tab:seqlab}
\end{center}
\end{table}

\noindent\textbf{Evaluation.}
Results for the text spans extraction models on manual matching test set are presented in Table~\ref{tab:seqlab}, which shows that CRF achieves the best performance and outperforms the neural models (LSTM-CRF, LM-LSTM-CRF). The LM-LSTM-CRF which contains character level language model is even worse (shown as w/ LM in table, with different character level hidden size). One reason could be the neural models require a large scale dataset to train, while our dataset does not meet this requirement. 

\subsection{Text Span-Pair Relation Extraction}
\noindent\textbf{Models.} For text span-pair relation extraction, we use the pre-trained BERT\textsubscript{BASE} model\footnote{https://github.com/huggingface/pytorch-pretrained-BERT} \cite{bert} as our backbone model to address the low-resource issue of our RE dataset generated from the limited CTA data. 
On this basis, we implement model variants of injecting context information awareness (Sec.~\ref{sec:context}) to utilize the full information in our dataset, which includes: hidden states Masking (Mask\textsubscript{AVG} and Mask\textsubscript{MAX}), Context position as Attention (C. Attn.) and Context postion as input Embedding (C. Emb.). For hidden states Masking, the different subscriptions represent different hidden state pooling methods (avg pooling and max pooling) For the two models using context position, we empirically use $E=30$ as the embedding size and truncate the context position sequence (Sec. \ref{sec:contextseq}) by $\pm10$. In addition, we experiment on the hierarchical BiLSTM model \cite{hbmp} and Piecewise Convolution Neural Network \cite{zeng2015distant} as the non-pretrained baseline models in comparison. Results are aggregated from 5 runs with different initialization seeds for all experiments.

\begin{table}[hbt!]
\small
\begin{center}
\begin{tabular}{ c | c c c }
\toprule
Sampling portion & Total & $<$next$>$ & $<$if$>$\\
\midrule
w/o sampling & 138670 & 693 & 131 \\
\midrule
6 : 3 : 1 & 1310 & 393 & 131 \\
4 : 2 : 1 & 917 & 262 & 131 \\
1 : 1 : 1 & 393 & 131 & 131 \\
\bottomrule
\end{tabular}
\caption{\textbf{Size of span-pair RE dataset}, by different sampling portion $<$none$> : <$next$> : <$if$>$}
\label{tab:dataset}
\end{center}
\end{table}
\noindent\textbf{Label sampling portion.}\label{sec:sampling}
The generated RE dataset has three types of label: $<$none$>$, $<$next$>$ and $<$if$>$, with a bias label distribution (Fig. \ref{tab:dataset}, w/o sampling). To leverage this, we do label sampling on the dataset. 

\noindent\textbf{Context level.} To capture the long-range context information useful to the CTA transcript parsing task, we use text spans with context $\textbf{t}_c$ (fig. \ref{fig:context}) as the input of models. The level of context is controlled by a hyperparameter $K$ (Fig. \ref{fig:context}). We experiment our models with different levels of context, while fixing the label sampling portion (Sec. \ref{sec:sampling}) to $<$none$>$ : $<$next$>$ : $<$if$> = 4 : 2 : 1$.

\begin{table*}[hbt!]
\small
\begin{center}
\begin{tabular}{ c | c c c c | c c c c }
\toprule
Setting & \multicolumn{4}{c|}{Generated Test Set} & \multicolumn{4}{c}{Manual Matching Test Set}\\
 & Accuracy & Micro F\textsubscript{1} & $<$next$>$ F\textsubscript{1} & $<$if$>$ F\textsubscript{1} & Accuracy & Micro F\textsubscript{1} & $<$next$>$ F\textsubscript{1} & $<$if$>$ F\textsubscript{1}\\
\midrule
BERT & 81.6 \textpm 1.0 & 70.1 \textpm 1.7 & 67.9 \textpm 3.2 & 73.4 \textpm 2.2 & 77.2 \textpm 2.7 & 62.2 \textpm 6.1 & 57.6 \textpm 6.4
& 72.4 \textpm 10.0\\
HBMP & 76.0 & 67.4 & - & - & 72.0 & 63.3 & - & - \\
PCNN & 58 & 40 & - & - & 56 & 43 & - & - \\
\midrule
C. Attn. & 82.5 \textpm 1.5 & 72.2 \textpm 2.6 & \textbf{70.9 \textpm 1.8} & 74.7 \textpm 4.4 & 81.2 \textpm 4.7 & 72.7 \textpm 7.5 & 68.7 \textpm
9.1 & \textbf{83.3 \textpm 5.5}\\
C. Emb. & \textbf{82.8 \textpm 1.4} & 72.7 \textpm 1.9 & 70.7 \textpm 2.8 & 76.3 \textpm 2.5 & 78.8 \textpm 8.5 & 67.4 \textpm 8.1 & 66.2 \textpm 9.2 & 67.5 \textpm 19.4\\
Mask\textsubscript{AVG} & 80.5 \textpm 2.7 & 69.0 \textpm 5.7 & 63.6 \textpm 7.1 & \textbf{77.0 \textpm 4.8} & 80.4 \textpm 7.1 & 73.4 \textpm 7.9 & 71.8 \textpm 9.4 & 79.2 \textpm 14.5\\
Mask\textsubscript{MAX} & 82.3 \textpm 1.4 & \textbf{72.6 \textpm 3.0} & 70.7 \textpm 3.2 & 76.1 \textpm 3.1 & \textbf{87.6 \textpm 1.5} & \textbf{81.4 \textpm 2.4} & \textbf{80.8 \textpm 1.9} & 83.3 \textpm 6.8\\
\bottomrule
\end{tabular}
\caption{\textbf{Performance of span-pair RE models,} with sampling portion $4 : 2 : 1$ and $K=2$. Evaluated on generated test set and manual matching test set.}
\label{tab:re}
\end{center}
\end{table*}
\noindent\textbf{Evaluation.} The results are available in Table \ref{tab:re}, which shows the model we proposed can outperform the baselines (BERT, HBMP, PCNN), and the model variant Mask\textsubscript{MAX} reach best performance among all variants when using context level $K = 2$ and sampling portion $= 4 : 2 : 1$. 
 
\begin{figure}[!bt]
 \centering
  \includegraphics[width=0.9\linewidth]{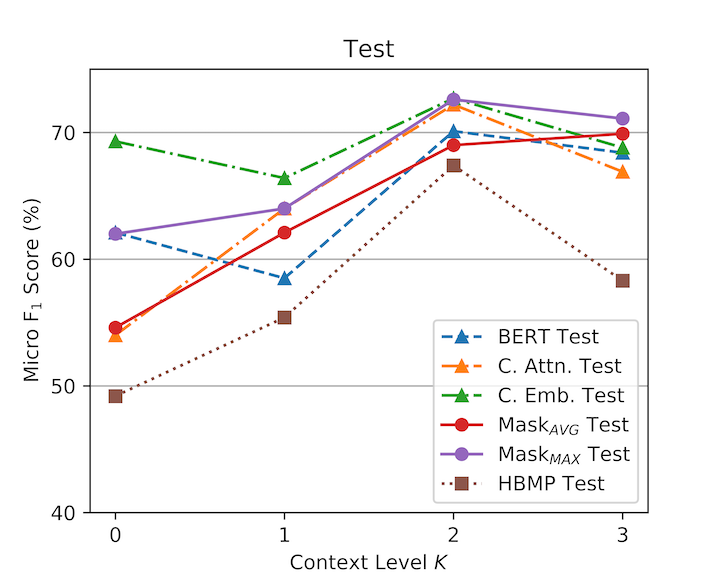}
 \caption{\textbf{The micro F\textsubscript{1} score of models on different context level $K$,} evaluated on generated test set.}
  \label{fig:curvecontexttest}
\end{figure}
\begin{table*}[hbt!]
\small
\begin{center}
\begin{tabular}{ c | c c c c | c c c c }
\toprule
Setting & \multicolumn{4}{c|}{Generated Test Set} & \multicolumn{4}{c}{Manual Matching Test Set}\\
 & Accuracy & Micro F\textsubscript{1} & $<$next$>$ F\textsubscript{1} & $<$if$>$ F\textsubscript{1} & Accuracy & Micro F\textsubscript{1} & $<$next$>$ F\textsubscript{1} & $<$if$>$ F\textsubscript{1}\\
\midrule
BERT \textsubscript{K=3} & 80.2 \textpm 3.2 & 68.4 \textpm 4.3 & 64.3 \textpm 6.6 & 74.6 \textpm 4.3 & 77.2 \textpm 3.0 & 63.6 \textpm 5.5 & 60.9 \textpm 7.2
& 70.3 \textpm 11.3\\
BERT \textsubscript{K=2} & 81.6 \textpm 1.0 & 70.1 \textpm 1.7 & 67.9 \textpm 3.2 & 73.4 \textpm 2.2 & 77.2 \textpm 2.7 & 62.2 \textpm 6.1 & 57.6 \textpm 6.4
& 72.4 \textpm 10.0\\
BERT \textsubscript{K=1} & 73.5 \textpm 2.7 & 58.5 \textpm 3.0 & 57.7 \textpm 4.7 & 60.0 \textpm 4.0 & 76.4 \textpm 2.3 & 60.2 \textpm 6.1 & 54.5 \textpm 7.0
& 76.1 \textpm 7.2\\
BERT \textsubscript{K=0} & 71.9 \textpm 2.6 & 62.1 \textpm 5.1 & 58.5 \textpm 6.9 & 69.0 \textpm 6.9 & 63.2 \textpm 5.2 & 50.7 \textpm 6.8 & 45.3 \textpm 10.2 & 71.0 \textpm 10.7\\
\midrule
Mask\textsubscript{MAX} \textsubscript{K=3} & 81.8 \textpm 0.9 & 71.1 \textpm 1.5 & 68.6 \textpm 1.9 & 75.3 \textpm 2.2 & 85.2 \textpm 4.1 & 78.6 \textpm 4.8 & 75.6 \textpm 6.1 & \textbf{88.9 \textpm 0.0}\\
Mask\textsubscript{MAX} \textsubscript{K=2} & \textbf{82.3 \textpm 1.4} & \textbf{72.6 \textpm 3.0} & \textbf{70.7 \textpm 3.2} & \textbf{76.1 \textpm 3.1} & \textbf{87.6 \textpm 1.5} & \textbf{81.4 \textpm 2.4} & \textbf{80.8 \textpm 1.9} & 83.3 \textpm 6.8\\
Mask\textsubscript{MAX} \textsubscript{K=1} & 76.3 \textpm 1.3 & 64.0 \textpm 1.9 & 58.4 \textpm 3.4 & 74.2 \textpm 1.6 & 69.2 \textpm 1.0 & 53.9 \textpm 1.5 & 47.6 \textpm 2.5 & 75.0 \textpm 0.0\\
Mask\textsubscript{MAX} \textsubscript{K=0} & 71.9 \textpm 2.7 & 62.0 \textpm 4.6 & 55.3 \textpm 7.5 & 73.9 \textpm 2.9 & 64.4 \textpm 2.7 & 52.7 \textpm 4.2 & 47.6 \textpm 5.6 & 71.7 \textpm 4.1\\
\bottomrule
\end{tabular}
\caption{\textbf{Performance of span-pair RE models on different context level $K$}, with sampling portion $4 : 2 : 1$. Evaluated on generated test set and manual matching test set.}
\label{tab:context}
\end{center}
\end{table*}
\begin{figure}[!bt]
 \centering
 \includegraphics[width=0.9\linewidth]{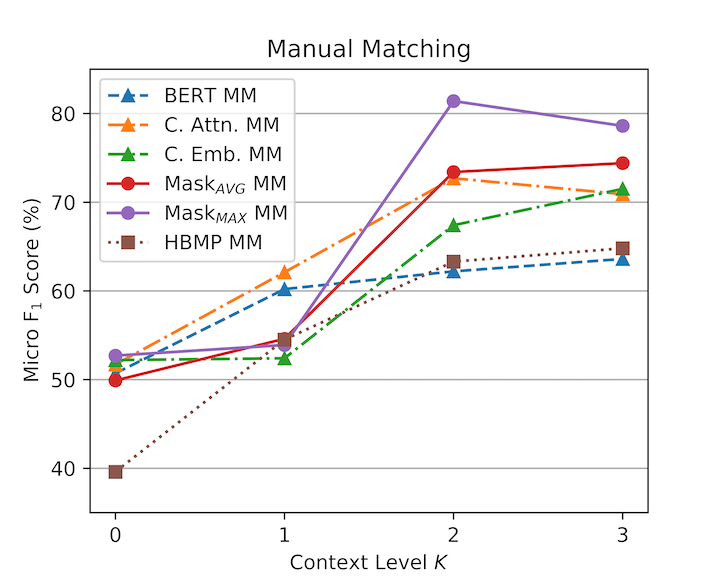}
 \caption{\textbf{The micro F\textsubscript{1} score of models on different context level $K$,} evaluated on manual matching test set.}
  \label{fig:curvecontextmanual}
\end{figure}

\noindent \textbf{Evaluation on context level.} The short version table of evaluation results for different context levels are shown in Table \ref{tab:context} and please refer to the appendix for the full version. The results are visualized in Fig.~\ref{fig:curvecontextmanual} and Fig.~\ref{fig:curvecontexttest}. The model Mask\textsubscript{MAX} reached the best micro F\textsubscript{1} score on the manual matching test set with context level $K=2$ over all models and $K$, which shows the effectiveness of the span-pair RE and the hidden state masking structure.

\begin{table}[hbt!]
\small
\begin{center}
\begin{tabular}{ c  c | c c }
\toprule
Portion & Model & \multicolumn{2}{c}{Micro F\textsubscript{1}} \\
 & & Generated & Manual \\
\midrule
$6 : 3 : 1$ & BERT & 67.6 \textpm 1.7 & 69.5 \textpm 4.6\\
& Mask\textsubscript{MAX} & 68.5 \textpm 2.1 & 71.1 \textpm 9.1 \\
\midrule
$4 : 2 : 1$ & BERT & 68.4 \textpm 4.3 & 63.6 \textpm 5.5 \\
& Mask\textsubscript{MAX} & 71.1 \textpm 1.5 & \textbf{78.6 \textpm 4.8} \\
\midrule
$1 : 1 :1$ & BERT & 65.4 \textpm 4.9 & 62.7 \textpm 3.4 \\
& Mask\textsubscript{MAX} & 70.3 \textpm 2.9 & 69.8 \textpm 3.5 \\
\bottomrule
\end{tabular}
\caption{\textbf{Performance on text spans relation extraction models on different label sampling settings,} with $K=3$. \textit{Generated} represent the sampled generated test set follows the sampling portion the model trained on, while \textit{Manual} represents the manual matching test set which is fixed to $6 : 3 : 1$.}
\label{tab:sampling}
 \vspace{-0.5cm}
\end{center}
\end{table}
\noindent \textbf{Evaluation on label sampling.}
We try 3 sampling settings and find $<$none$> : <$next$> : <$if$> = 4 : 2 : 1$ shows the best performance on manual matching test set for most cases (Table \ref{tab:sampling}). Please refer to the appendix for the full results on label sampling.

\noindent\textbf{Discussion.}
We have some observations when looking through the results on manual matching test set: (1) The model variants injected with context information awareness are more sensitive to the change of context level $K$, comparing to the vanilla BERT model. These variants are outperforming the vanilla model when provided with more context, but would fall behind if provided with short even no context. (2) Vanilla models without specific context awareness structures (BERT, HBMP, PCNN) also gain improvements from the context on the manual matching test set. (3) A big gap of $<$next$>$ F\textsubscript{1} score between $K=1$ and $K=2$ are observed in most of the models. This is because when $K=1$ context only provide the sentence enclosing the text span, the $K=2$ context is providing the last and the next sentence, which is useful for predicting the $<$next$>$ relation.

The results on generated test set (Fig.\ref{fig:curvecontexttest}) is also interesting, in which the performance is not stably increased as the $K$ increasing. This may be caused by the propagation of error from fuzzy matching. Since there are some error (noisy) samples in the generated dataset, the models are more likely to capture the noisy patterns from the noisy samples. The larger the context is, the more noisy patterns are contained. Still, changing $K$ from $K=1$ to $K=2$ gives noticeable improvement to all models, especially for the $<$next$>$ F\textsubscript{1} score.

Also, the experiments on label sampling (Table \ref{tab:sampling}, see appendix for the full result) show the performance of models are sensitive to sampling portion. Resampling and reweighting techniques for alleviating label imbalance could be helpful to address such problem in future study.
\section{Conclusion}
In this paper, we explored automated CTA transcript parsing, which is a challenging task due to the lack of direct supervision data and the requirement of document level understanding. We proposed a weakly supervised framework to utilize the full information in data. We noticed the importance of context in the CTA parsing task and exploited model variants to make use of context information. Our evaluation on manually labeled test set shows the effectiveness of our framework.

\section{Acknowledgment}
This work has been supported in part by National Science Foundation SMA 18-29268, Schmidt Family Foundation, Amazon Faculty Award, Google Research Award, and JP Morgan AI Research Award. We would like to thank all the collaborators in INK research lab for their constructive feedback on the work. We thank the anonymous reviewers for their valuable feedback.


\bibliography{acl2019}
\bibliographystyle{acl_natbib}
\newpage
\appendix
\label{sec:appendix}
\newpage
\begin{table*}[htb!]
\begin{center}
\begin{tabular}{ | c | l | c | }
\hline
\# & Protocol Phrase & Matching status \\
 & Matched Text Span & \\
\hline
1 & with your non-dominant hand   & over  \\ 
  & And now with your non-dominant hand you &\\
\hline
2 & Pass wire   & correct \\ 
  & pass wire &\\
\hline
3 & resistance   & noisy  \\ 
  & -  &\\
\hline
4 & remove the wire by bringing it back into the housing   & correct \\ 
  & remove the wire , bring it back into the little housing &\\
\hline
5 & Put syringe back on and confirm that there is still blood flow   & miss  \\ 
  & syringe back on and make sure you still have good flow &\\
\hline
6 & Remove needle  & correct \\
 & remove your needle & \\
\hline
7 & leave wire in place   & correct \\ 
  & Leave the wire in place &\\
\hline
8 & Make a nick in the skin that is wide enough for the catheter   & correct \\ 
  & make a nick in the skin wide enough for whatever catheter &\\
\hline
9 & Pass dilator   & correct \\ 
  & pass dilator &\\
\hline
10 & Remove dilator   & miss  \\ 
  & out dilator &\\
\hline
11 & while holding the wire in place   & correct \\ 
  & leaving the wire in place , always holding onto the wire &\\
\hline
12 & Put catheter through the wire   & correct \\ 
  & putting the catheter through the wire &\\
\hline
13 & Remove wire   & correct \\ 
  & remove the wire &\\
\hline
14 & Check and irrigate all ports   & wrong  \\ 
  & So we have task one , decide on location &\\
\hline
15 & Lock the catheter   & correct \\ 
  & lock the catheter &\\
\hline
16 & attach with Luer-lock   & wrong \\ 
  & lock the catheter &\\
\hline
17 & Suture in place   & wrong \\ 
  & Task five , insert needle &\\
\hline
18 & Verify placement with x-ray   & miss \\ 
  & Verify placement with &\\
\hline
19 & Prepare patient   & correct \\ 
  & prep the patient &\\
\hline
20 & self & noisy \\
 & -  &\\
\hline
\end{tabular}
\caption{\textbf{\textbf{Case study for text span matching,} using Glove-300d as the sentence encoding method.}}
\label{tab:fullcontext}
\end{center}
\end{table*}

\begin{table*}[h!]
\small
\begin{center}
\begin{tabular}{ c | c c c c | c c c c }
\toprule
Setting & \multicolumn{4}{c|}{Generated Test Set} & \multicolumn{4}{c}{Manual Matching Test Set}\\
 & Accuracy & Micro F\textsubscript{1} & $<$next$>$ F\textsubscript{1} & $<$if$>$ F\textsubscript{1} & Accuracy & Micro F\textsubscript{1} & $<$next$>$ F\textsubscript{1} & $<$if$>$ F\textsubscript{1}\\
\midrule
BERT \textsubscript{K=3} & 80.2 \textpm 3.2 & 68.4 \textpm 4.3 & 64.3 \textpm 6.6 & 74.6 \textpm 4.3 & 77.2 \textpm 3.0 & 63.6 \textpm 5.5 & 60.9 \textpm 7.2
& 70.3 \textpm 11.3\\
BERT \textsubscript{K=2} & 81.6 \textpm 1.0 & 70.1 \textpm 1.7 & 67.9 \textpm 3.2 & 73.4 \textpm 2.2 & 77.2 \textpm 2.7 & 62.2 \textpm 6.1 & 57.6 \textpm 6.4
& 72.4 \textpm 10.0\\
BERT \textsubscript{K=1} & 73.5 \textpm 2.7 & 58.5 \textpm 3.0 & 57.7 \textpm 4.7 & 60.0 \textpm 4.0 & 76.4 \textpm 2.3 & 60.2 \textpm 6.1 & 54.5 \textpm 7.0
& 76.1 \textpm 7.2\\
BERT \textsubscript{K=0} & 71.9 \textpm 2.6 & 62.1 \textpm 5.1 & 58.5 \textpm 6.9 & 69.0 \textpm 6.9 & 63.2 \textpm 5.2 & 50.7 \textpm 6.8 & 45.3 \textpm 10.2 & 71.0 \textpm 10.7\\
\midrule
C. Attn. \textsubscript{K=3} & 80.2 \textpm 3.7 & 66.9 \textpm 5.6 & 66.1 \textpm 6.9 & 68.0 \textpm 5.4 & 81.6 \textpm 4.1 & 70.9 \textpm 8.1 & 67.9 \textpm
10.1 & 79.8 \textpm 5.1\\
C. Attn. \textsubscript{K=2} & 82.5 \textpm 1.5 & 72.2 \textpm 2.6 & 70.9 \textpm 1.8 & 74.7 \textpm 4.4 & 81.2 \textpm 4.7 & 72.7 \textpm 7.5 & 68.7 \textpm
9.1 & 83.3 \textpm 5.5\\
C. Attn. \textsubscript{K=1} & 75.4 \textpm 2.1 & 64.0 \textpm 3.8 & 58.8 \textpm 4.6 & 73.5 \textpm 2.5 & 74.0 \textpm 2.5 & 62.1 \textpm 2.5 & 53.1 \textpm
4.0 & 86.6 \textpm 8.6\\
C. Attn. \textsubscript{K=0} & 67.3 \textpm 3.1 & 54.0 \textpm 4.6 & 43.7 \textpm 7.5 & 72.5 \textpm 2.3 & 58.8 \textpm 3.7 & 51.7 \textpm 4.0 & 43.6 \textpm
2.9 & 83.3 \textpm 6.8\\
\midrule
C. Emb. \textsubscript{K=3} & 79.8 \textpm 2.5 & 68.8 \textpm 4.0 & 64.0 \textpm 4.6 & 76.9 \textpm 3.8 & 80.4 \textpm 7.1 & 71.5 \textpm 10.1 & 68.4 \textpm
9.4 & 81.2 \textpm 12.6\\
C. Emb. \textsubscript{K=2} & \textbf{82.8 \textpm 1.4} & \textbf{72.7 \textpm 1.9} & \textbf{70.7 \textpm 2.8} & 76.3 \textpm 2.5 & 78.8 \textpm 8.5 & 67.4 \textpm 8.1 & 66.2 \textpm 9.2 & 67.5 \textpm 19.4\\
C. Emb. \textsubscript{K=1} & 76.5 \textpm 2.3 & 66.4 \textpm 3.5 & 62.1 \textpm 3.5 & 74.4 \textpm 5.0 & 67.6 \textpm 7.7 & 52.4 \textpm 10.7 & 43.3 \textpm
11.7 & 79.8 \textpm 6.4\\
C. Emb. \textsubscript{K=0} & 77.5 \textpm 1.2 & 69.3 \textpm 2.9 & 63.9 \textpm 3.0 & \textbf{78.4 \textpm 6.0} & 67.6 \textpm 6.6 & 52.2 \textpm 6.6 & 40.7 \textpm 5.5 & 83.7 \textpm 7.2\\
\midrule
Mask\textsubscript{AVG} \textsubscript{K=3} & 81.4 \textpm 1.3 & 69.9 \textpm 3.4 & 67.6 \textpm 2.6 & 73.6 \textpm 6.3 & 81.6 \textpm 3.2 & 74.4 \textpm 7.2 & 71.0 \textpm 8.1 & 86.1 \textpm 5.6\\
Mask\textsubscript{AVG} \textsubscript{K=2} & 80.5 \textpm 2.7 & 69.0 \textpm 5.7 & 63.6 \textpm 7.1 & 77.0 \textpm 4.8 & 80.4 \textpm 7.1 & 73.4 \textpm 7.9 & 71.8 \textpm 9.4 & 79.2 \textpm 14.5\\
Mask\textsubscript{AVG} \textsubscript{K=1} & 74.7 \textpm 1.2 & 62.1 \textpm 2.1 & 55.9 \textpm 2.4 & 73.2 \textpm 2.4 & 71.2 \textpm 3.2 & 54.6 \textpm 4.4 & 46.8 \textpm 5.1 & 77.8 \textpm 5.6\\
Mask\textsubscript{AVG} \textsubscript{K=0} & 67.1 \textpm 2.2 & 54.6 \textpm 3.2 & 45.7 \textpm 5.5 & 71.1 \textpm 2.0 & 59.2 \textpm 4.1 & 49.9 \textpm 3.6 & 42.0 \textpm 4.8 & 80.6 \textpm 6.8\\
\midrule
Mask\textsubscript{MAX} \textsubscript{K=3} & 81.8 \textpm 0.9 & 71.1 \textpm 1.5 & 68.6 \textpm 1.9 & 75.3 \textpm 2.2 & 85.2 \textpm 4.1 & 78.6 \textpm 4.8 & 75.6 \textpm 6.1 & \textbf{88.9 \textpm 0.0}\\
Mask\textsubscript{MAX} \textsubscript{K=2} & 82.3 \textpm 1.4 & 72.6 \textpm 3.0 & 70.7 \textpm 3.2 & 76.1 \textpm 3.1 & \textbf{87.6 \textpm 1.5} & \textbf{81.4 \textpm 2.4} & \textbf{80.8 \textpm 1.9} & 83.3 \textpm 6.8\\
Mask\textsubscript{MAX} \textsubscript{K=1} & 76.3 \textpm 1.3 & 64.0 \textpm 1.9 & 58.4 \textpm 3.4 & 74.2 \textpm 1.6 & 69.2 \textpm 1.0 & 53.9 \textpm 1.5 & 47.6 \textpm 2.5 & 75.0 \textpm 0.0\\
Mask\textsubscript{MAX} \textsubscript{K=0} & 71.9 \textpm 2.7 & 62.0 \textpm 4.6 & 55.3 \textpm 7.5 & 73.9 \textpm 2.9 & 64.4 \textpm 2.7 & 52.7 \textpm 4.2 & 47.6 \textpm 5.6 & 71.7 \textpm 4.1\\
\midrule
HBMP \textsubscript{K=3} & 68.0 & 58.3 & - & - & 74.0 & 64.8 & - & - \\
HBMP \textsubscript{K=2} & 76.0 & 67.4 & - & - & 72.0 & 63.3 & - & - \\
HBMP \textsubscript{K=1} & 67.0 & 55.4 & - & - & 60.0 & 54.5 & - & - \\
HBMP \textsubscript{K=0} & 50.0 & 49.2 & - & - & 50.0 & 39.6 & - & - \\
\midrule
PCNN \textsubscript{K=3} & 47 & 31 & - & - & 62 & 48 & - & - \\
PCNN \textsubscript{K=2} & 58 & 40 & - & - & 56 & 43 & - & - \\
PCNN \textsubscript{K=1} & 44 & 28 & - & - & 50 & 28 & - & - \\
PCNN \textsubscript{K=0} & 44 & 29 & - & - & 34 & 24 & - & - \\
\bottomrule
\end{tabular}
\caption{\textbf{Performance of text spans relation extraction models on different context level $K$}, with sampling portion $4 : 2 : 1$.}
\label{tab:fullcontext}
\end{center}
\end{table*}

\begin{table*}[h!]
\small
\begin{center}
\begin{tabular}{ c | c c c c | c c c c }
\toprule
Model & \multicolumn{4}{c|}{Sampled Generated Test Set} & \multicolumn{4}{c}{Manual Matching Test Set}\\
 & Accuracy & Micro F\textsubscript{1} & $<$next$>$ F\textsubscript{1} & $<$if$>$ F\textsubscript{1} & Accuracy & Micro F\textsubscript{1} & $<$next$>$ F\textsubscript{1} & $<$if$>$ F\textsubscript{1}\\
\midrule
\multicolumn{9}{c}{Sampling portion = $6 : 3 : 1$ (1.3k samples)}\\
\midrule
BERT & 79.0 \textpm 1.2 & 67.6 \textpm 1.7 & 68.3 \textpm 1.3 & 65.5 \textpm 3.2 & 80.0 \textpm 2.5 & 69.5 \textpm 4.6 & 72.2 \textpm 1.8 & 52.0 \textpm 31.1\\
C. Attn. & 75.6 \textpm 2.4 & 61.4 \textpm 4.0 & 62.9 \textpm 4.0 & 57.4 \textpm 6.0 & 80.4 \textpm 4.3 & 68.8 \textpm 7.4 & 66.4 \textpm 8.9 & 75.2 \textpm 10.4\\
C. Emb. & 77.9 \textpm 1.8 & 65.7 \textpm 1.2 & 66.4 \textpm 2.3 & 64.0 \textpm 4.1 & 80.8 \textpm 2.0 & 70.7 \textpm 4.1 & 70.0 \textpm 5.4 & 71.8 \textpm 8.8\\
Mask\textsubscript{AVG} & 79.8 \textpm 1.0 & 69.1 \textpm 2.4 & 68.7 \textpm 2.2 & 70.4 \textpm 3.4 & 81.6 \textpm 3.4 & 72.3 \textpm 6.5 & 69.0 \textpm 6.9 & 83.3 \textpm 6.8\\
Mask\textsubscript{MAX} & 80.1 \textpm 0.8 & 68.5 \textpm 2.1 & \textbf{69.5 \textpm 2.8} & 65.9 \textpm 1.9 & 81.6 \textpm 5.0 & 71.1 \textpm 9.1 & 66.9 \textpm 11.6 & 83.3 \textpm 6.8\\
\midrule
\multicolumn{9}{c}{Sampling portion = $4 : 2 : 1$ (0.9k samples)}\\
\midrule
BERT & 80.2 \textpm 3.2 & 68.4 \textpm 4.3 & 64.3 \textpm 6.6 & 74.6 \textpm 4.3 & 77.2 \textpm 3.0 & 63.6 \textpm 5.5 & 60.9 \textpm 7.2 & 70.3 \textpm 11.3\\
C. Attn. & 80.2 \textpm 3.7 & 66.9 \textpm 5.6 & 66.1 \textpm 6.9 & 68.0 \textpm 5.4 & 81.6 \textpm 4.1 & 70.9 \textpm 8.1 & 67.9 \textpm 10.1 & 79.8 \textpm 5.1\\
C. Emb. & 79.8 \textpm 2.5 & 68.8 \textpm 4.0 & 64.0 \textpm 4.6 & 76.9 \textpm 3.8 & 80.4 \textpm 7.1 & 71.5 \textpm 10.1 & 68.4 \textpm 9.4 & 81.2 \textpm 12.6\\
Mask\textsubscript{AVG} & 81.4 \textpm 1.3 & 69.9 \textpm 3.4 & 67.6 \textpm 2.6 & 73.6 \textpm 6.3 & 81.6 \textpm 3.2 & 74.4 \textpm 7.2 & 71.0 \textpm 8.1 & 86.1 \textpm 5.6\\
Mask\textsubscript{MAX} & \textbf{81.8 \textpm 0.9} & \textbf{71.1 \textpm 1.5} & 68.6 \textpm 1.9 & 75.3 \textpm 2.2 & \textbf{85.2 \textpm 4.1} & \textbf{78.6 \textpm 4.8} & \textbf{75.6 \textpm 6.1} & \textbf{88.9 \textpm 0.0}\\
\midrule
\multicolumn{9}{c}{Sampling portion = $1 : 1 : 1$ (0.4k samples)}\\
\midrule
BERT & 64.6 \textpm 4.6 & 65.4 \textpm 4.9 & 51.3 \textpm 3.5 & 79.3 \textpm 6.0 & 69.6 \textpm 5.6 & 62.7 \textpm 3.4 & 54.3 \textpm 3.6 & 87.3 \textpm 7.3\\
Mask\textsubscript{AVG} & 64.6 \textpm 2.9 & 64.2 \textpm 3.4 & 46.7 \textpm 7.4 & 79.6 \textpm 1.7 & 72.8 \textpm 1.6 & 63.5 \textpm 2.3 & 57.4 \textpm 2.3 & 83.6 \textpm 4.4\\
Mask\textsubscript{MAX} & 68.8 \textpm 3.5 & 70.3 \textpm 2.9 & 55.6 \textpm 4.6 & \textbf{83.9 \textpm 1.2} & 77.6 \textpm 3.2 & 69.8 \textpm 3.5 & 63.2 \textpm 2.4 & 88.4 \textpm 7.6\\
\bottomrule
\end{tabular}
\caption{\textbf{Performance on text spans RE models on different label sampling settings,} with $K=3$. Sampled generated test set follows the sampling portion the model trained on while manual matching test set is fixed.}
\label{tab:fullsampling}
\end{center}
\end{table*}

\end{document}